# A Data-Oriented Approach to Semantic Interpretation[1]

Rens Bod*, Remko Bonnema* and Remko Scha*

**Abstract**

In Data-Oriented Parsing (DOP), an annotated language corpus is used as a stochastic grammar. The most probable analysis of a new input sentence is constructed by combining sub-analyses from the corpus in the most probable way. This approach has been succesfully used for syntactic analysis, using corpora with syntactic annotations such as the Penn Treebank. If a corpus with semantically annotated sentences is used, the same approach can also generate the most probable semantic interpretation of an input sentence. The present paper explains this semantic interpretation method, and summarizes the results of a preliminary experiment. Semantic annotations were added to the syntactic annotations of most of the sentences of the ATIS corpus. A data-oriented semantic interpretation algorithm was succesfully tested on this semantically enriched corpus.

## 1    Introduction

The Data-Oriented Parsing (DOP) method suggested in Scha (1990) and developed in Bod (1992-1995) is a probabilistic parsing strategy which does not single out a narrowly predefined set of structures as the statistically significant ones. It accomplishes this by maintaining a large corpus of analyses of previously occurring utterances. New input is analyzed by combining arbitrarily large subtrees from the corpus in the most probable way.

DOP is motivated by cognitive considerations. Psychological experiments indicate that the frequency with which various structures have occurred in a person's past language experience creates strong biases when ambiguous new input is analyzed (e.g. Pearlmutter & MacDonald, 1992; Mitchell et al., 1992; Juliano & Tanenhaus, 1993; Gibson & Loomis, 1994). Data-Oriented Parsing accounts for this by taking the frequencies of occurrence of corpus-subtrees into account when estimating the likelihood of different alternative analyses, and by then choosing the most probable one.

For the syntactic dimension of language understanding, various instantiations of this Data-Oriented approach have been worked out (e.g. Bod, 1992-1996; Sima'an, 1995-1996). A method for extending it to the semantic domain was first suggested by (van den Berg et al., 1994). In the present paper we discuss a computationally effective version of that method, and an implemented system that used it.

We first summarize the Data-Oriented Parsing method now. Then we show that this method can be straightforwardly extended into a semantic analysis method, if a corpus is used in which the trees are enriched with semantic annotations. Finally, we report on an experiment with a semantically augmented version of the ATIS corpus (Marcus et al., 1993).

---

[1] This work was partially supported by the Netherlands Organization for Scientific Research (Priority Programme Language and Speech Technology).

* Institute for Logic, Language and Computation, University of Amsterdam, Spuistraat 134, 1012 VB Amsterdam, The Netherlands, Rens.Bod@let.uva.nl, Bonnema@mars.let.uva.nl, Remko.Scha@let.uva.nl

## 2 Corpus-Driven Syntactic Analysis

So far, the data-oriented parsing method has been worked out for (and tested on) corpora with simple syntactic annotations, consisting of labelled trees. Let us illustrate this with a very simple imaginary example. Suppose that a corpus consists of only two trees:

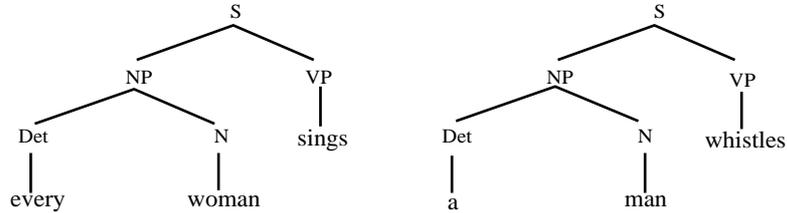

Figure 1. Imaginary corpus of two trees.

We employ one operation for combining subtrees, called "composition", indicated as ∘; this operation identifies the leftmost nonterminal leaf node of one tree with the root node of a second tree (i.e., the second tree is *substituted* on the leftmost nonterminal leaf node of the first tree). A new input sentence like "*A woman whistles*" can now be parsed by combining subtrees from this corpus[2]. For instance:

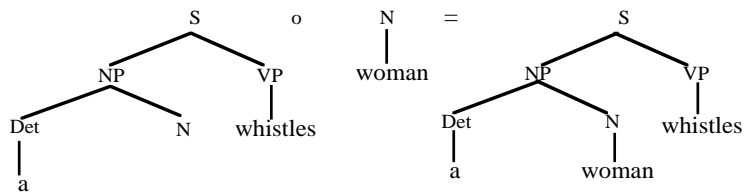

Figure 2. Derivation and parse for "*A woman whistles*"

Other derivations may yield the same parse tree; for instance[3]:

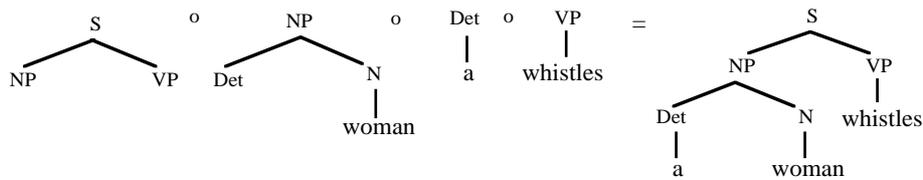

Figure 3. Different derivation generating the same parse for "*A woman whistles*"

or

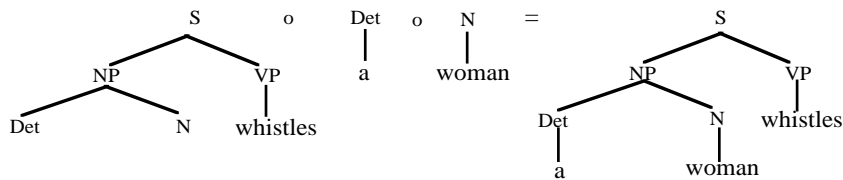

Figure 4. Another derivation generating the same parse for "*A woman whistles*"

---

[2] In this simple example, the sentence "*a woman whistles*" is not ambiguous.

[3] Here $t \circ u \circ v \circ w$ should be read as $((t \circ u) \circ v) \circ w$.

Thus, a parse tree can have many derivations involving different corpus-subtrees. DOP estimates the probability of substituting a subtree $t_i$ on a specific node as the probability of selecting $t_i$ among all subtrees in the corpus that could be substituted on that node. This probability may be estimated as the number of occurrences of a subtree $t_i$, divided by the total number of occurrences of subtrees $t$ with the same root node label as $t_i$: $\#(t_i) / \#(t : \mathrm{root}(t) = \mathrm{root}(t_i))$. The probability of a derivation $t_1 \circ ... \circ t_n$ can be computed as the product of the probabilities of the substitutions that it involves: $\Pi_i \#(t_i) / \#(t : \mathrm{root}(t) = \mathrm{root}(t_i))$. The probability of a parse tree is equal to the probability that any of its derivations occurs, which is the sum of the probabilities of all derivations of that parse tree. If a parse tree has $k$ derivations: $(t_{11} \circ ... \circ t_{1j} \circ ...), ..., (t_{i1} \circ ... \circ t_{ij} \circ ...), ..., (t_{k1} \circ ... \circ t_{kj} \circ ...)$, its probability can be written as: $\Sigma_i \Pi_j \#(t_{ij}) / \#(t : \mathrm{root}(t) = \mathrm{root}(t_{ij}))$.

The DOP method distinguishes itself from other statistical approaches, such as Pereira and Schabes (1992), Black et al. (1993) and Briscoe (1994), in that it does not predefine or train a formal grammar, but it takes subtrees directly from hand-parsed sentences in a treebank with a probability proportional to the number of occurrences of these subtrees in the treebank.

Bod (1993a) shows that DOP can be implemented using conventional context-free parsing techniques. To select the most probable parse, Bod (1993b) gives a Monte Carlo approximation algorithm. In Sima'an (1995), an efficient polynomial algorithm for a sub-optimal solution is given. The model was tested on the Air Travel Information System (ATIS) corpus as analyzed in the Penn Treebank (Marcus et al., 1993), achieving better test results than other stochastic grammars (cf. Bod, 1996; Sima'an 1996). On Penn's Wall Street Journal corpus, the data-oriented parsing approach has been tested by Sekine & Grishman (1995), and by Charniak (1996). Though Charniak only uses corpus-subtrees smaller than depth 2 (which in our experience constitutes a less-than-optimal version of the data-oriented parsing method), he reports that his program "outperforms all other non-word-based statistical parsers/grammars on this corpus".

## 3  Corpus-Driven Semantic Analysis

To use the Data-Oriented Parsing method not just for syntactic analysis, but also for semantic interpretation, four steps must be taken:
(1) decide on a formalism for representing the meanings of sentences and surface-constituents.
(2) annotate the corpus-sentences and their surface-constituents with such semantic representations.
(3) establish a method for deriving the meaning representations associated with arbitrary corpus-subtrees and with compositions of such subtrees.
(4) reconsider the probability calculations.
We now discuss these four steps.

**(1) Semantic formalism**

The decision about the representational formalism is to some extent arbitrary, as long as it has a well-defined model-theory and is rich enough for representing the meanings of sentences and constituents that are relevant for the intended application domain. For our exposition in this paper we will use a wellknown standard formalism: extensional type theory (see Gamut, 1991), i.e., a higher-order logical language that combines lambda-abstraction with connectives and quantifiers. The only currently implemented system for data-oriented semantic interpretation (Bonnema, 1996) uses a different logical language, however. And in many application contexts it probably makes sense to use an A.I.-style language which highlights domain structure (frames, slots, and fillers), while limiting the use of quantification and negation.

**(2) Semantic annotation**

We assume a corpus that is already syntactically annotated as before: with labelled trees that indicate surface constituent structure. Now our basic idea, taken from van den Berg et al. (1994), is to augment this syntactic annotation with a semantic one: to every meaningful syntactic node, we add a type-logical formula that expresses the meaning of the corresponding surface-constituent. If we would carry out this idea in a completely direct way, the toy corpus of Figure 1 might, for instance, turn into the toy corpus of Figure 4.

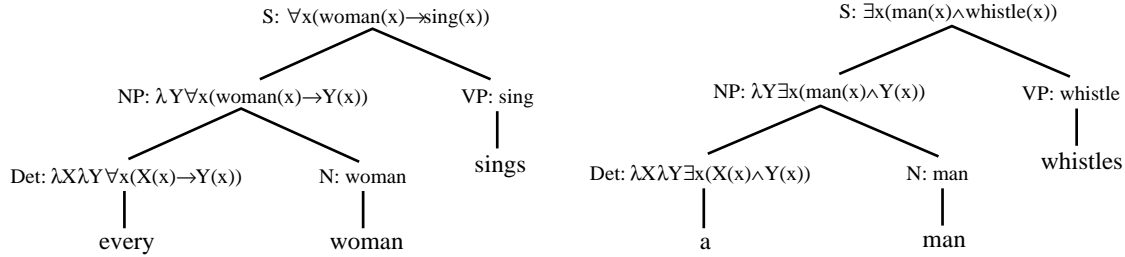

Figure 4. Imaginary corpus of two trees with syntactic and semantic labels.

Van den Berg et al. indicate how a corpus of this sort may be used for data-oriented semantic interpretation. Their algorithm, however, requires a procedure which can inspect the semantic formula of a node and determine the contribution of the semantics of a lower node, in order to be able to "factor out" that contribution. The details of this procedure have not been specified. We can avoid the need for this procedure by using an annotation which indicates explicitly how the semantic formula for a node is built up on the basis of the semantic formulas of its daughter nodes.

We therefore indicate the semantic annotation of the corpus trees as follows. (1) For every meaningful lexical node a type logical formula is specified that represents its meaning. (2) For every meaningful non-lexical node a formula schema is specified which indicates how its meaning representation may be put together out of the formulas assigned to its daughter nodes. In the examples below, these schemata use the variable *d1* to indicate the meaning of the leftmost daughter constituent, *d2* to indicate the meaning of the second daughter constituent, etc. Using this notation, the semantically annotated version of the toy corpus of Figure 1 is the toy corpus rendered in Figure 5.

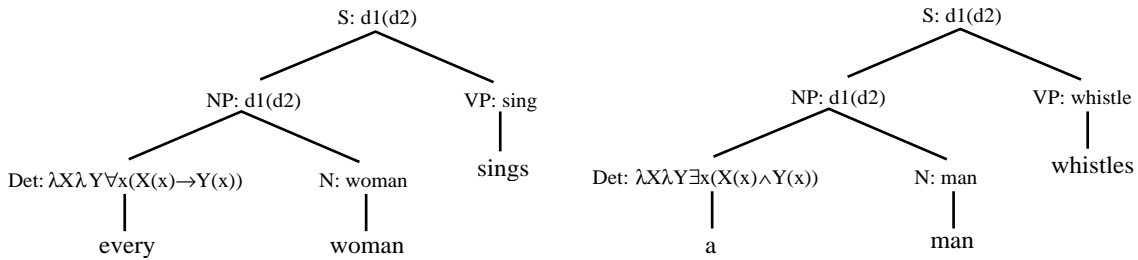

Figure 5. Same imaginary corpus of two trees with syntactic and semantic labels using the daughter notation

This kind of semantic annotation is what will be used in the experiment described in section 4 of this paper. It may be noted that the rather oblique description of the semantics of the higher nodes in the tree would easily lead to mistakes, if annotation would be carried out completely manually. An annotation tool that makes the expanded versions of the formulas visible for the annotator is obviously called for. Such a tool was developed by Bonnema (1996).

This annotation convention obviously assumes that the meaning representation of a surface-constituent *can* in fact always be composed out of the meaning representations of its subconstituents. This assumption is not unproblematic. To maintain it in the face of phenomena such as non-standard quantifier scope or "discontinuous constituents" creates complications in the syntactic or semantic analyses assigned to certain sentences and their constituents. It is therefore not clear yet whether our current treatment ought to be viewed as completely general, or whether a treatment in the vein of van den Berg et al. (1994) should be worked out.

**(3) The meanings of subtrees and their compositions**

As in the purely syntactic version of DOP, we now want to compute the probability of a (semantic) analysis by considering all the different ways in which it can be generated by combining subtrees from the corpus. We can do this in virtually the same way. The only novelty is a slight modification in the process by which a corpus tree is decomposed into subtrees, and a corresponding modification in the composition operation which combines subtrees. If we extract a subtree out of a tree, we replace the semantics of the new leaf node with a unification variable of the same type. Correspondingly, when the composition operation substitutes a subtree at this node, this unification variable is unified with the semantic formula on the substituting tree. (It is required that the semantic type of this formula matches the semantic type of the unification variable.)

A simple example will make this clear. First, let us consider what subtrees the corpus makes available now. For instance, one of the decompositions of the annotated corpus sentence "*A man whistles*" is the following:

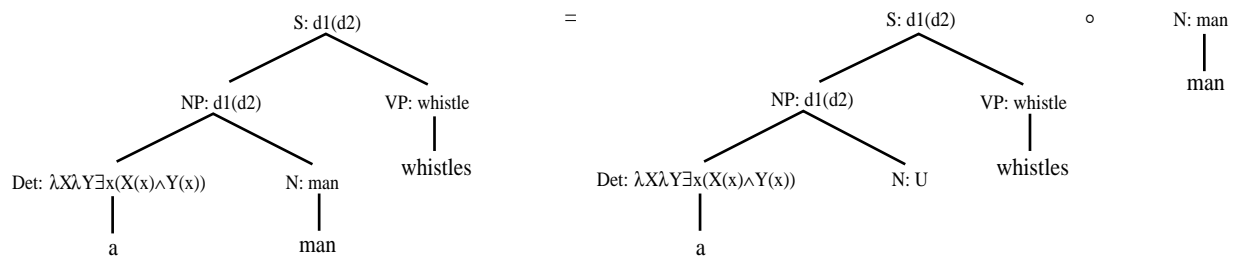

Figure 6. Decomposing a tree into subtrees with unification variables

We see that by decomposing the tree into two subtrees, the semantics at the breakpoint-node *N: man* is replaced by a variable. Now an analysis for the sentence "*a woman whistles*" can, for instance, be generated in the following way.

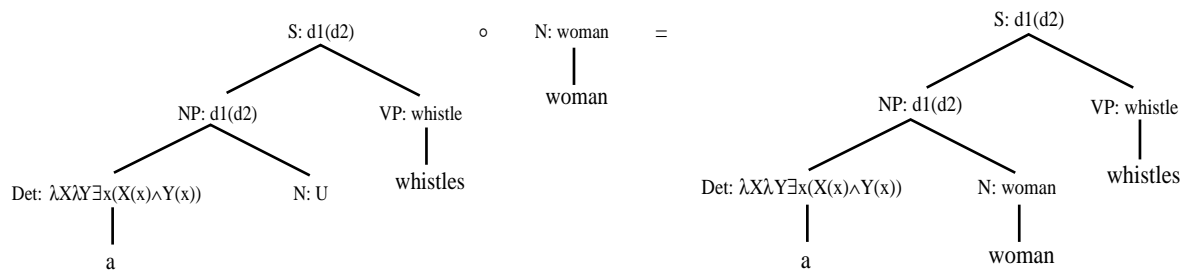

Figure 7. Generating an analysis for "*A woman whistles*".

**(4) The Statistical Model of Data-Oriented Semantic Interpretation**

We now define the probability of an interpretation of an input string, given a partially annotated corpus.

Given a partially annotated corpus as defined above, the multiset of corpus subtrees consists of all subtrees with a well-defined top-node semantics, that are generated by applying to the trees of the corpus the decomposition mechanism described above. The probability of substituting a certain subtree on a leaf node of category *C* is the probability of selecting this subtree among all subtrees with category *C* in the multiset. This probability is equal to the ratio between the number of occurrences of this subtree and the total number of occurrences of subtrees with category *C*.

A derivation of a string is a tuple of subtrees, such that their composition results in a tree whose yield is the string. The probability of a derivation is the product of the substitution probabilities of these subtrees.

A tree resulting from a derivation of a string is called an analysis of this string. The probability of an analysis is the probability that any of its derivations occurs; this is the sum of the probabilities of all its derivations.

An interpretation of a string is a formula which is provably equivalent to the semantic annotation of the top node of an analysis of this string. The probability of a string $s$ with an interpretation $I$ is the sum of the probabilities of the analyses of string $s$ with a top node annotated with a formula that is provably equivalent to $I$.

## 4      An application to the ATIS domain

A first experiment with the ideas sketched above was carried out by Bonnema (1996), who developed a semantically annotated corpus on the basis of the ATIS corpus in the Penn Treebank (Marcus et al., 1993). We summarize his results here.

**The semantic formalism**

To provide the ATIS corpus with semantic annotations, Bonnema used a higher-order logical language that differs in certain ways from the standard higher-order logic we used in the examples above. In the Air Travel Information System application, most of the nouns in most of the sentences intuitively correspond to *functions* rather than to sets of individuals. It does not make sense to analyze the noun "*arrival-time*" as denoting a set of timepoints, for instance. It must be analyzed as denoting a function which yields a timepoint when it is applied to a flight. Most of the ATIS sublanguage partakes in this phenomenon. For instance, flights also have fares, departure-times, and destinations. There are also two-place functions, such as flights, distances and travel-times which take pairs of cities as arguments. Maintaining the traditional linguistic semantic representations for sentences involving functional nouns is not impossible, but it does create complexities that one may want to avoid (cf. De Bruin and Scha, 1988). For this reason, Bonnema used non-standard semantic annotations, and a new logical language with additional set-theoretical operations, that facilitate representing the meanings of complex sentences with plural function-nouns ("*the fares of the flights from Boston to the West Coast cities*").

This logical language was called LLARQ: "Logical Language for Aviation Related Questions". This makes it sound more ad hoc than it is. Only one aspect of LLARQ *is* ad hoc: its descriptive constants were chosen in such a way that they cover the Air Travel Information System application in a reasonable way. Going through the 750 sentences of the ATIS corpus, 34 occasions were encountered where the meaning of a sentence was completely unclear, or where the representation of only that sentence would have required the introduction of one or more special descriptive constants in the logic. These 34 sentences were discarded, leaving a corpus of 716 sentences.

The syntactic annotations of these 716 sentences were obtained from the Penn Treebank -- also when the annotations found there seemed inconsistent or otherwise non-optimal. The syntactic trees were then extended with LLARQ annotations, using the method discussed in the previous sections. Figures 8 and 9 give an impression of the resulting annotations.

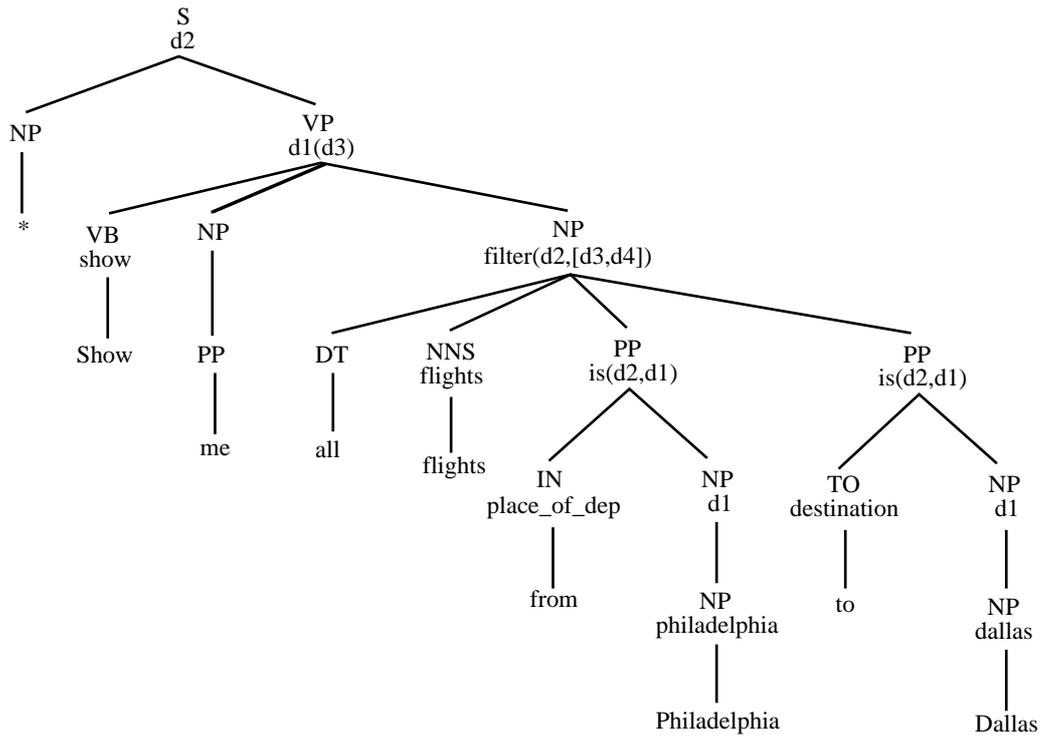

Figure 8. Annotation for the ATIS sentence "*Show me all flights from Philadelphia to Dallas*"

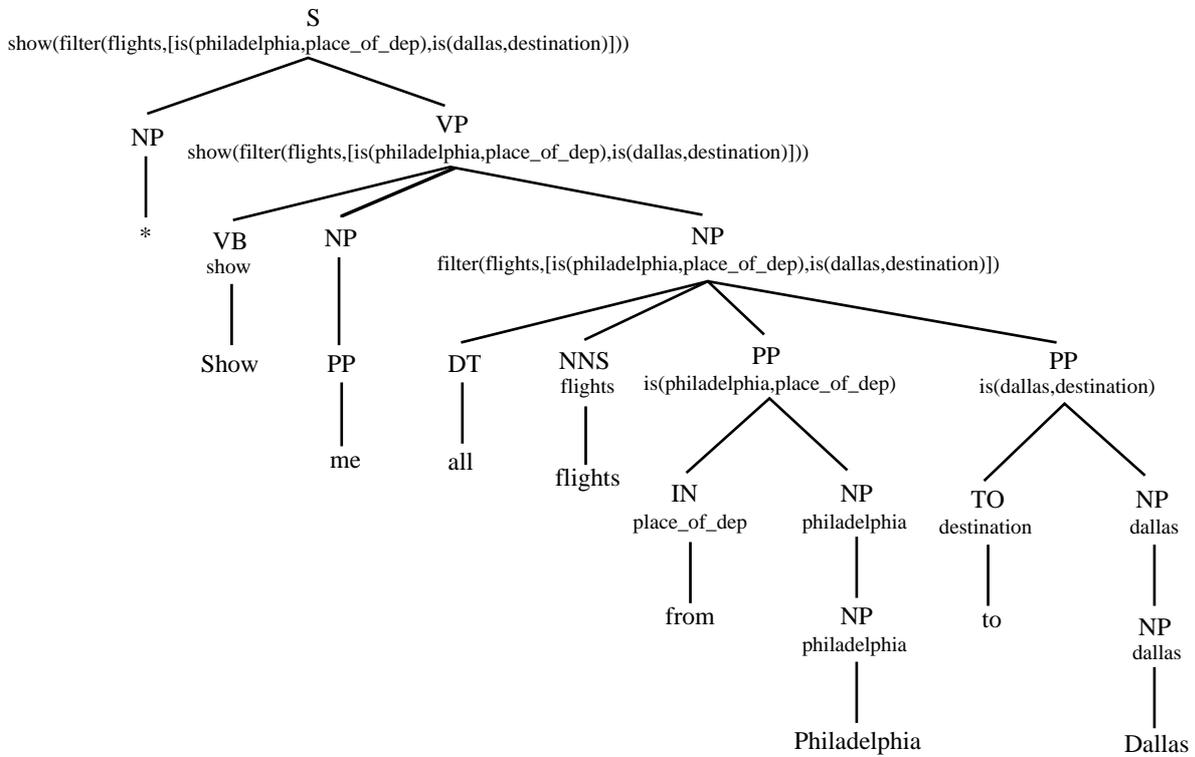

Figure 9. Annotation for the same sentence with expanded versions of the formulas

**The annotation task: SEMTAGS**

The semantic annotation of the 716 ATIS trees may seem a laborious and error-prone process. In order to expedite this task, Bonnema created a flexible and powerful tool (SEMTAGS) for semi-automatically enriching trees with semantic annotations. SEMTAGS is a graphical interface, written in C, using the XVIEW toolkit. All annotated trees can be evaluated and checked by a Prolog interpreter that runs in the background.

SEMTAGS incrementally creates a first order Markov model based on existing annotations. In a fraction of a second (on a Sparc 2) a proposal for the semantic annotation of a new syntactic tree is given. The annotator may then check or fill in the missing annotations by hand; subsequently, SEMTAGS checks the type consistency of the annotation.

After the first 100 sentences of the corpus had been annotated and checked by hand, SEMTAGS already produced the correct annotations for 80% of the nodes for the immediately subsequent sentences. This number increased considerably as more sentences were annotated. Although the probabilistic mechanism of SEMTAGS is far from being adequate as a general model for statistical interpretation, it gives reasonable annotation proposals very fast, and removes the need for tedious copying of annotations.

**Experimental evaluation**

Having annotated, checked and double-checked the 716 ATIS trees, Bonnema divided the corpus into a 666 sentence training set and a 50 sentence test set. The division was random except for one constraint: that all words in the test set actually occurred in the training set. The training set analyses were converted into subtrees (depth ≤ 10) with substitution probabilities. These subtrees served as the stochastic tree-grammar used by the DOP parser (Sima'an, 1996). For the 50 test sentences, the parse results were as follows:

* 28 matched exactly (semantically and syntactically)
* 31 obtained the correct syntactic structure (though the semantics was not necessarily correct)
* 44 obtained the correct semantics (i.e. a semantic top node formula logically equivalent to the formula in the test set, though the syntactic structure did not necessarily match exactly)
* 5 obtained incorrect semantics and incorrect syntactic structure
* 1 could not be parsed

Although these experiments are still preliminary, they seem to indicate that semantic parsing is quite robust, even if the syntactic analysis is incorrect. This can be concuded from the fact that for 44 sentences the correct semantic interpretations were generated, while for 13 of these the syntactic structure was "incorrect" (i.e. did not match with the test set structure).

This robustness may be explained by the fact that the syntactic annotations in the Penn Treebank are often not consistent. For instance, NPs such as "*all the flights*" are annotated sometimes as a binary structure and sometimes as a ternary structure. However, their semantic annotations at the NP-level are equivalent, regardless their underlying structure. Since we are actually interested in the correct semantic interpretation of a whole utterance (e.g. for interfacing with databases), and not so much in the correct semantic interpretations of all subconstituents, we conclude that semantic analysis is relatively insensitive to syntactic annotation errors.

It may also be interesting to mention that *without* semantic annotations, only 27 parses matched exactly with the test set trees (compared to 31 parses in case semantic annotations were included). Thus, semantic annotations also lead to better *syntactic* results. This may be explained by the fact that semantic annotations have a restrictive effect, thus leading to fewer ambiguities. This semantic restrictiveness also contributed to better time performance: *with* semantic annotations the 50 test sentences were parsed 6 times as fast as *without* semantic annotations.

## Acknowledgements

We are grateful to Khalil Sima'an for using his DOP parser and for useful discussions.